\pdfoutput=1

\documentclass[11pt]{article}

\usepackage{emnlp2021}

\usepackage{times}
\usepackage{latexsym}

\usepackage[T1]{fontenc}

\usepackage[utf8]{inputenc}

\usepackage{microtype}

\usepackage{array}
\usepackage{multirow}
\usepackage{makecell}
\usepackage{soul}
\usepackage{url}
\usepackage{graphicx}
\usepackage{amsmath}
\usepackage{booktabs}
\usepackage{algorithm}
\usepackage{subfigure}
\usepackage{algorithmic}
\usepackage{enumitem}
\usepackage{amsthm,amsmath,amssymb}
\usepackage{mathrsfs}
\usepackage{bm}
\usepackage{hyperref}

\makeatletter
\newcommand{\thickhline}{%
	\noalign {\ifnum 0=`}\fi \hrule height 1pt
	\futurelet \reserved@a \@xhline
}

\title{Self- and Pseudo-self-supervised Prediction of Speaker and Key-utterance\\ for Multi-party Dialogue Reading Comprehension}

\author{Yiyang Li$^{1,2,3}$ \and Hai Zhao$^{1,2,3,*}$ \\
	$^1$ Department of Computer Science and Engineering, Shanghai Jiao Tong University\\
	$^2$ Key Laboratory of Shanghai Education Commission for Intelligent Interaction\\and Cognitive Engineering, Shanghai Jiao Tong University\\
	$^3$ MoE Key Lab of Artificial Intelligence, AI Institute, Shanghai Jiao Tong University\\
	\texttt{eric-lee@sjtu.edu.cn,zhaohai@cs.sjtu.edu.cn}
}

\begin{document}
	\maketitle
	\begin{abstract}
		Multi-party dialogue machine reading comprehension (MRC) brings tremendous challenge since it involves multiple speakers at one dialogue, resulting in intricate speaker information flows and noisy dialogue contexts. To alleviate such difficulties, previous models focus on how to incorporate these information using complex graph-based modules and additional manually labeled data, which is usually rare in real scenarios. In this paper, we design two labour-free self- and pseudo-self-supervised prediction tasks on speaker and key-utterance to implicitly model the speaker information flows, and capture salient clues in a long dialogue. Experimental results on two benchmark datasets have justified the effectiveness of our method over competitive baselines and current state-of-the-art models.
	\end{abstract}
	
	\let\thefootnote\relax\footnotetext{*Corresponding author. This paper was partially supported by Key Projects of National Natural Science Foundation of China (U1836222 and 61733011).}
	
	\section{Introduction}
	Dialogue machine reading comprehension (MRC, \citealp{hermann2015teaching}) aims to teach machines to understand dialogue contexts so that solves multiple downstream tasks (\citealp{yang2019friendsqa}; \citealp{li2020molweni}; \citealp{lowe2015ubuntu}; \citealp{wu2017sequential}; \citealp{zhang2018modeling}). In this paper, we focus on question answering (QA) over dialogue, which tests the capability of a model to understand a dialogue by asking it questions with respect to the dialogue context. QA over dialogue is of more challenge than QA over plain text (\citealp{rajpurkar2016squad}; \citealp{reddy2019coqa}; \citealp{yang2019friendsqa}) owing to the fact that conversations are full of informal, colloquial expressions and discontinuous semantics. Among this, multi-party dialogue brings even more tremendous challenge compared to two-party dialogue (\citealp{sun2019dream}; \citealp{cui2020mutual}) since it involves multiple speakers at one dialogue, resulting in complicated discourse structure \citep{li2020molweni} and intricate speaker information flows. Besides this, \citet{zhang2021multi} also pointed that for long dialogue contexts, not all utterances contribute to the final answer prediction since a lot of them are noisy and carry no useful information.
	
\begin{figure}[tbp]
		\includegraphics[width=0.45\textwidth]{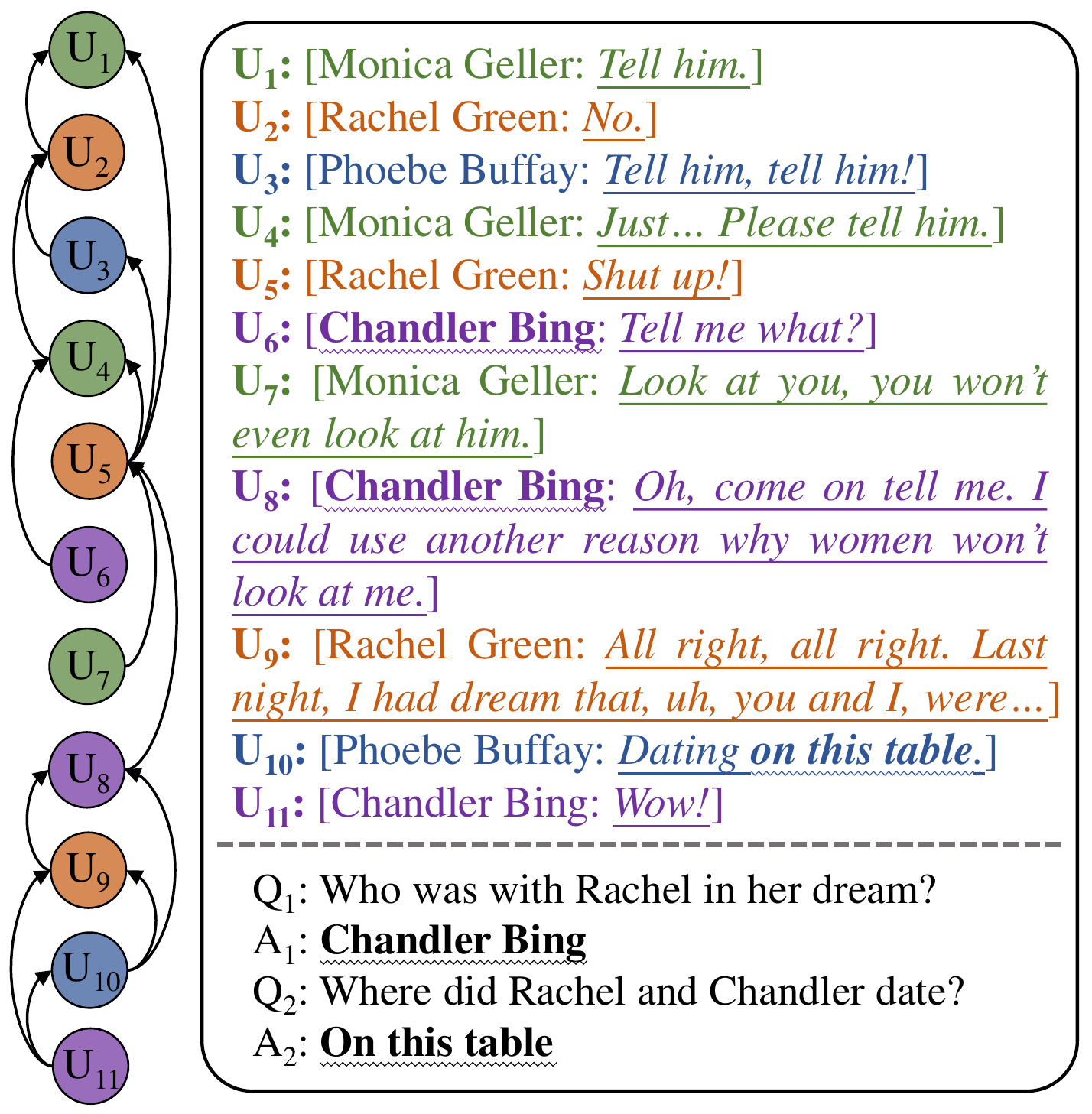}
		\centering
		\caption{Right part: A dialogue and its corresponding questions from FriendsQA, whose answers are marked with wavy lines. Left part: The speaker information flows of this dialogue.} 
		\label{speaker_info_flow}
	\end{figure}

	To illustrate the challenge of multi-party dialogue MRC, we extract a dialogue example from FriendsQA dataset \citep{yang2019friendsqa} which is shown in Figure \ref{speaker_info_flow}.
	This single dialogue involves four different speakers with intricate speaker information flows. The arrows here represent the direction of information flows, from senders to receivers. Let us consider the reasoning process of $\rm Q_1$: a model should first notice that it is \emph{Rachel} who had a dream and locate $\rm U_9$, then solve the coreference resolution problem that \emph{I} refers to \emph{Rachel} and \emph{you} refers to \emph{Chandler}. This coreference knowledge must be obtained by considering the information flow from $\rm U_9$ to $\rm U_8$, which means \emph{Rachel} speaks to \emph{Chandler}. $\rm Q_2$ follows a similar process, a model should be aware of that $\rm U_{10}$ is a continuation of $\rm U_9$ and solves the above coreference resolution problem as well.
	
	To tackle the aforementioned obstacles, we design a self-supervised speaker prediction task to implicitly model the speaker information flows, and a pseudo-self-supervised key-utterance prediction task to capture salient utterances in a long and noisy dialogue. In detail, the self-supervised speaker prediction task guides a carefully designed Speaker Information Decoupling Block (SIDB, introduced in Section \ref{SIDB}) to decouple speaker-aware information, and the key-utterance prediction task guides a Key-utterance Information Decoupling Block (KIDB, introduced in Section \ref{KIDB}) to decouple key-utterance-aware information. We finally fuse these two kinds of information and make final span prediction to get the answer of a question.
	
	To sum up, the main contributions of our method are three folds:
	\begin{itemize}[leftmargin=*, topsep=1pt]
		\setlength{\itemsep}{0pt}
		\setlength{\parsep}{0pt}
		\setlength{\parskip}{0pt}
		\item We design a novel self-supervised speaker prediction task to better capture the indispensable speaker information flows in multi-party dialogue. Compared to previous models, our method requires no additional manually labeled data which is usually rare in real scenarios.
		\item We design a novel key-utterance prediction task to capture key-utterance information in a long dialogue context and filter noisy utterances.
		\item Experimental results on two benchmark datasets show that our model outperforms strong baselines by a large margin, and reaches comparable results to the current state-of-the-art models even under the condition that they utilized additional labeled data.
	\end{itemize}
	
	\section{Related work}
	\subsection{Pre-trained Language Models}
	Recently, pre-trained language models (PrLMs), like BERT \citep{devlin2019bert}, RoBERTa \citep{liu2019roberta}, ALBERT \citep{lan2019albert}, XLNet \citep{yang2019xlnet} and ELECTRA \citep{clark2020electra}, have reached remarkable achievements in learning universal natural language representations by pre-training large language models on massive general corpus and fine-tuning them on downstream tasks (\citealp{socher2013recursive}; \citealp{wang2018glue}; \citealp{wang2019superglue}; \citealp{lai2017race}). We argue that the self-attention mechanism \citep{vaswani2017attention} in PrLMs is in essence a variant of Graph Attention Network (GAT, \citealp{velivckovic2017graph}), which has an intrinsic capability of exchanging information. Compared to vanilla GAT, a Transformer block consisting of residual connection \citep{He_2016_CVPR} and layer normalization \citep{ba2016layer} is more stable in training. Hence, it is chosen as the basic architecture of our SIDB (Section \ref{SIDB}) and KIDB (Section \ref{KIDB}) instead of vanilla GAT.
	
	\subsection{Multi-party Dialogue Modeling}
	There are several previous works that study multi-party dialogue modeling on different downstream tasks such as response selection and dialogue emotion recognition. \citet{ijcai2019-696} utilize the \emph{response to (@)} labels and a Graph Neural Network (GNN) to explicitly model the speaker information flows. \citet{wang2020response} design a pre-training task named Topic Prediction to equip PrLMs with the ability of tracking parallel topics in a multi-party dialogue. \citet{jia2020multi} make use of an additional labeled dataset to train a dependency parser, then utilize the dependency parser to disentangle parallel threads in multi-party dialogues. \citet{ghosal2019dialoguegcn} propose a window-based heterogeneous Graph Convolutional Network (GCN) to model the emotion flow in multi-party dialogues.
	
	\subsection{Speaker Information Incorporation}
	In dialogue MRC, speaker information plays a significant role in comprehending the dialogue context. In the latest studies, \citet{liu2021filling} propose a Mask-based Decoupling-Fusing Network (MDFN) to decouple speaker information from dialogue contexts, by adding inter-speaker and intra-speaker masks to the self-attention blocks of Transformer layers. However, their approach is restricted to two-party dialogue since they have to specify the sender and receiver roles of each utterance. \citet{gu2020speaker} propose Speaker-Aware BERT (SA-BERT) to capture speaker information by adding speaker embedding at token representation stage of the Transformer architecture, then pre-train the model using next sentence prediction (NSP) and masked language model (MLM) losses. Nonetheless, their speaker embedding lacks of well-designed pre-training task to refine, resulting in inadequate speaker-specific information. Different from previous models, our model is suitable for the more challenging multi-party dialogue and is equipped with carefully-designed task to better capture the speaker information.
	
	\section{Methodology}
\begin{figure*}[htbp]
		\includegraphics[width=1.0\textwidth]{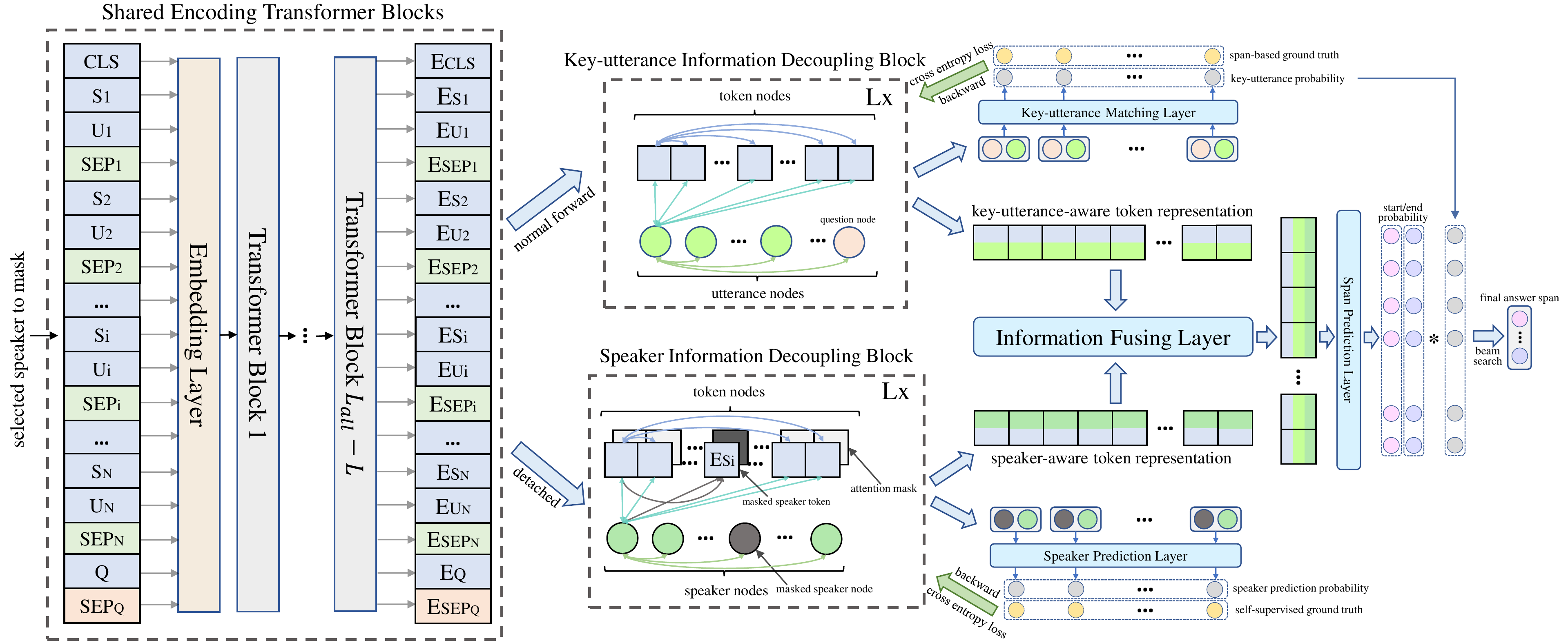}
		\centering
		\caption{The overview of our model, which contains a shared Transformer encoder, a key-utterance information decoupling block, a speaker information decoupling block and a fusion-prediction layer. In speaker information decoupling block, the bi-directional arrow means that the information flows from and to both sides, the unidirectional arrow means that the information only flows from start nodes to end nodes.} 
		\label{overview}
	\end{figure*}

	In this part, we will formulate our task and present our proposed model as shown in Figure \ref{overview}. There are four main parts in our model, a shared Transformer encoder, a key-utterance information decoupling block, a speaker information decoupling block and a final fusion-prediction layer. In the following sections, we will introduce these modules in detail.
	
	\subsection{Task Formulation}
	\label{Task_Formulation}
	Let $\mathbb{C} = \{U_1, U_2, ..., U_N\}$ be a dialogue context with $N$ utterances. Each utterance $U_i = \{S_i, W_i\}$ consists of a speaker $S_i$ specified by a name and a sequence of words $W_i$ speaker $S_i$ utters. $W_i$ can be denoted as a $l_i$-length sequence $\{w_{i1}, w_{i2}, ..., w_{i_{li}}\}$. Let a question corresponds to the dialogue context be $\mathbb{Q} = \{q_1, q_2, ..., q_L\}$, where $L$ is the length of the question and each $q_i$ is a token of the question. Given $\mathbb{C}$ and $\mathbb{Q}$, a dialogue MRC model is required to find an answer $a$ for the question, which is restricted to be a continuous span of the dialogue context. In some datasets, $a$ can be an empty string indicating that there is no answer to the question according to the dialogue context.
	
	\subsection{Shared Transformer Encoder}
	\label{STE}
	To fully utilize the powerful representational ability of PrLMs, we employ a \emph{pack} and \emph{separate} method as \citet{zhang2021multi}, which is supposed to take advantage of the deep Transformer blocks to make the context and question better interacted with each other. We first pack the context and question as a joint input to feed into the Transformer blocks and separate them according to the position for further interaction.
	
	Given the dialogue context $\mathbb{C}$ and a corresponding question $\mathbb{Q}$, we pack them to form a sequence:\\
	$\mathbb{X}$ = \{[CLS]$\rm \mathbb{Q}$[SEP]$\rm S_1$:$\rm U_1$[SEP]$\dots$
	$\rm S_N$:$\rm U_N$ [SEP]\},
	where [CLS] and [SEP] are two special tokens and each $\rm S_i$:$\rm U_i$ pair is the name and utterance of a speaker separated by a colon. This sequence $\mathbb{X}$ is then fed into $L_{all}-L$ layers of Transformer blocks to gain its contextualized representation $\bm{E}\in \mathcal{R}^{J\times d}$ where $J$ is the length of the sequence after tokenized by Byte-Pair Encoding (BPE) tokenizer \citep{sennrich2016neural} and $d$ is the hidden dimension of the Transformer block. Here $L_{all}$ is the total number of Transformer layers specified by the type of the PrLM, $L$ is a hyper-parameter which means the number of decoupling layers.
	
	\subsection{Key-utterance Information Decoupling Block}
	\label{KIDB}
	Given the contextualized representation $\bm{E}$ from Section \ref{STE}, follow \citet{zhang2021multi}, we gather the representation of [SEP] tokens from $\bm{E}$ as the representation of each utterance in the dialogue context. These representations are used to initialize $N$ utterance nodes $\bm{E_U}=\{\bm{E_{u_i}}\in \mathcal{R}^d\}_{i=1}^{N}$ and a question node $\bm{E_q}\in \mathcal{R}^d$ as illustrated in the middle-upper part of Figure \ref{overview}. The representations of normal tokens are gathered as token nodes $\bm{E_T}=\{\bm{E_{t_i}}\in \mathcal{R}^d\}_{i=1}^{n}$ where $n$ is the number of normal tokens in the dialogue context. Then, another $L$ layers of multi-head self-attention Transformer blocks are used to exchange information inter- and intra- the three types of nodes:
	\begin{equation}
		\label{eq1}
		\begin{aligned}
			& {\rm Attn}(Q, K, V) = {\rm softmax}(\frac{QK^T}{\sqrt{d_k}})V\\
			& {\rm head_i} = {\rm Attn}(EW_i^Q, EW_i^K, EW_i^V)\\
			& {\rm MultiHead}(E) = [{\rm head}_1,\dots,{\rm head}_h]W^O
		\end{aligned}
	\end{equation}
	Here $W_i^Q\in \mathcal{R}^{d\times d_q}$, $W_i^K\in \mathcal{R}^{d\times d_k}$, $W_i^V\in \mathcal{R}^{d\times d_v}$, $W^O\in \mathcal{R}^{hd_v\times d}$ are matrices with trainable weights, $h$ is the number of attention heads and $[;]$ denotes the concatenation operation.
	
	After stacking $L$ layers of multi-head self-attention: ${\rm MultiHead}([\bm{E_U};\bm{E_q};\bm{E_T}])$ to fully exchange information between these nodes, we get a question representation $\bm{H_q}\in \mathcal{R}^d$, the utterance representations $\bm{H_U} = \{\bm{H_{u_i}}\in \mathcal{R}^d\}_{i=1}^N$, and the token representations $\bm{H_T} = \{\bm{H_{t_i}}\in \mathcal{R}^d\}_{i=1}^n$.
	
	$\bm{H_q}$ is then paired with each $\bm{H_{u_i}}$ to conduct the key-utterance prediction task. In detail, we use a heuristic matching mechanism proposed by \citep{mou2016natural} to calculate the matching score of the question representation and utterance representation. Here we define a matching function $Match(\bm{X}, \bm{Y}, \rm {activ})$, where $\bm{X}, \bm{Y} \in \mathcal{R}^{d*N}$, as follows:
	\begin{equation}
		\label{match_func}
		\begin{aligned}
			&\bm{G} = [\bm{X}; \bm{Y}; \bm{X}-\bm{Y}; \bm{X}\odot \bm{Y}] \in \mathcal{R}^{4d\times N}\\
			&\bm{P} = {\rm activ}(\bm{a}^T\bm{G}) \in \mathcal{R}^N
		\end{aligned}
	\end{equation}
	Here $\odot$ denotes element-wise multiplication and $\bm{a}\in \mathcal{R}^{4d}$ is a vector with trainable weights. The $\rm activ$ is an activation function to get a probability distribution according to the downstream loss function, which can be chosen from $softmax$ and $sigmoid$. In span-based dialogue MRC datasets, we set the pseudo-self-supervised key-utterance target based on the position of the answer span. We name it pseudo-self-supervised since it is generated from the original span labels, but requires no additional labeled data. Specifically, we set $p^{target} = i$ where $i$ is the index of the utterance that contains the answer span. Then we calculate the key-utterance distribution by:
	\begin{equation}
		\begin{aligned}
			& \bm{H_Q} = \{\bm{H_q}\}_{i=1}^N \in \mathcal{R}^{d\times N}\\
			& \bm{P_U}^{pred} = Match(\bm{H_U}, \bm{H_Q}, softmax)
		\end{aligned}
	\end{equation}
	$\bm{P_U}^{pred}\in \mathcal{R}^N$ is later expanded to the length of token nodes to get $\bm{P_U}^{expand}\in \mathcal{R}^n$ which will be put forward to filter noisy utterances in the fusion-prediction layer (introduce in Section \ref{FPL}). We adopt cross-entropy loss to compute the loss of this task:
	\begin{equation}
		\mathcal{L}_U = -log(\bm{P_U}^{pred}[p^{target}])
	\end{equation}
	The gradient of $\mathcal{L}_U$ will flow backwards to refine the representations of the utterance nodes so that they can decouple key-utterance-aware information from the original representations. After the interaction between token nodes and utterance nodes, the token nodes will gather key-utterance-aware information from the utterance nodes. Therefore, we denote the token representations as key-utterance-aware: $\bm{H_T}^k = \bm{H_T}\in \mathcal{R}^{d\times n}$, which will be forwarded to the fusion-prediction layer described in Section \ref{FPL}.
	
	\subsection{Speaker Information Decoupling Block}
	\label{SIDB}
	This part is the core of our model, which contributes to modeling the complex speaker information flows. In this section, we first introduce the self-supervised speaker prediction task we proposed, then depict the decoupling process of speaker information.
	\subsubsection{Self-supervised Speaker Prediction}
	As defined in Section \ref{Task_Formulation}, we have a dialogue context $\mathbb{C} = \{U_1, U_2, ..., U_N\}$ where each utterance $U_i = \{S_i, W_i\}$ consists of a speaker $S_i$ specified by a name. We randomly choose an $\rm m_{th}$ utterance and mask its speaker name. Then for every $(U_i, U_m)$ pair where $i \neq m$, the model should determine whether they are uttered by the same speaker, that is to say, whether $S_i = S_m$.
	
\begin{figure}[htbp]
		\includegraphics[width=0.45\textwidth]{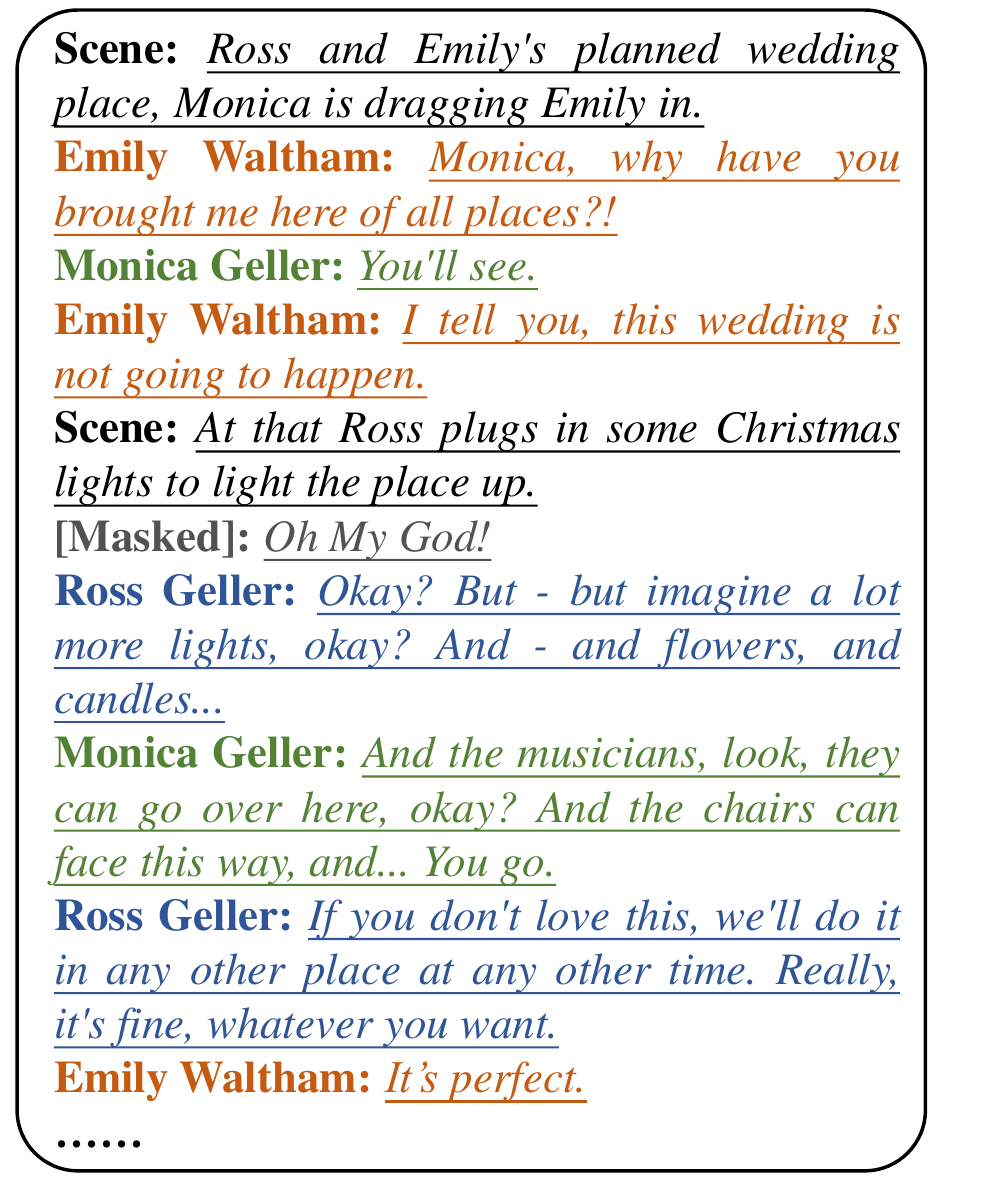}
		\centering
		\caption{An example of the speaker prediction task, which involves three speakers. \emph{Scene} here is a narrative description which introduces some additional information about the scene.}
		\label{speaker_prediction_exp}
	\end{figure}

	We figure this task a relatively difficult one since it requires the model to have a thorough understanding of the speaker information flows and solve problems such as coreference resolution. Figure \ref{speaker_prediction_exp} is an example of the self-supervised speaker prediction task, where the speaker of the utterance in gray is masked. We human can determine that the masked speaker should be \emph{Emily Waltham} by considering that \emph{Ross} and \emph{Monica} is persuading \emph{Emily} to attend the wedding by showing her the wedding place, and when \emph{Monica} and \emph{Emily} reaches there, it should be \emph{Emily} who is surprised to say \emph{"Oh My God"}. However, it is not that easy for machines to capture these information flows.
	
	\subsubsection{Speaker Information Decoupling}
	To fully utilize the interactive feature of self-attention mechanism \citep{vaswani2017attention} and the powerful representational ability of PrLMs, we also use Transformer blocks to capture the interactive speaker information flows and fulfill this difficult task.
	
	We first detach $\bm{E}$ from the computational graph to get $\bm{E^{de}}$, then as what we do in Section \ref{KIDB}, the representation of [SEP] tokens are gathered from $\bm{E^{de}}$ to initialize $N-1$ unmasked speaker nodes $\bm{E_S}=\{\bm{E_{s_i}}\in \mathcal{R}^d\}_{i=1}^{N-1}$ and a masked speaker node $\bm{E_{s_m}}\in \mathcal{R}^d$. The representation of normal tokens are gathered as token nodes $\bm{E_T}=\{\bm{E_{t_i}}\in \mathcal{R}^d\}_{i=1}^{n}$. Then, we add attention mask to the token nodes corresponding to the selected speaker name before they are forwarded into the speaker information decoupling block, as illustrated in the middle-lower part of Figure \ref{overview}. The reasons why we use this detach-mask strategy are as follows. First, we mask the selected speaker before the speaker information decoupling block instead of at the very beginning before the encoder since it is better to let the utterance decoupling block see all the speaker names. Based on this point, we detach $\bm{E}$ from the computational graph and add attention mask to avoid target leakage. If we use a normal forward instead, the encoder would simply attend to the speaker names, which would hurt performance (discuss in detail in Section \ref{detaching}). Besides, this strategy also helps the model better decouple the key-utterance-aware and speaker-aware information from the original representations.
	
	In detail, the mask strategy is similar as \citet{liu2021filling}. We modify Eq. (\ref{eq1}) to:
	\begin{equation}
		\label{eq5}
		\begin{aligned}
			& {\rm Attn}(Q, K, V, M) = {\rm softmax}(\frac{QK^T}{\sqrt{d_k}}+M)V\\
			& {\rm head_i} = {\rm Attn}(EW_i^Q, EW_i^K, EW_i^V, M)\\
			& {\rm MultiHead}(E, M) = [{\rm head}_1,\dots,{\rm head}_h]W^O
		\end{aligned}
	\end{equation}
	Let the start index and end index of the masked speaker tokens be $m_s$ and $m_e$, to make the selected speaker name unseen to other nodes, the attention mask is obtained as follows:
	\begin{equation}
		\bm{M_S}[i, j]=\left\{
		\begin{array}{cl}
			-\infty, & \text {if}\ j \in [m_s, m_e] \\
			0, & \text {otherwise}
		\end{array}\right.
	\end{equation}
	By adding this mask, other nodes will not attend to the masked token nodes, thus preventing target leakage. On the mean time, the speaker nodes will have to collect clues from other nodes through deep interaction to make prediction, which implicitly models the complex speaker information flows.
	
	After stacking $L$ layers of masked multi-head self-attention:  MultiHead([$\bm{E_S};\bm{E_{s_m}};\bm{E_T}],\bm{M_S}$]), we get a masked speaker representation $\bm{H_{s_m}}\in \mathcal{R}^d$, the normal speaker representation $\bm{H_S}=\{\bm{H_{s_i}}\in \mathcal{R}^d\}_{i=1}^{N-1}$, and the token representation $\bm{H_T} = \{\bm{H_{t_i}}\in \mathcal{R}^d\}_{i=1}^n$.
	
	$\bm{H_{s_m}}$ is then paired with each $\bm{H_{s_i}}$ to conduct the self-supervised speaker prediction task. We also adopt the matching function defined in Eq. (\ref{match_func}):
	\begin{equation}
		\begin{aligned}
			& \bm{H_M} = \{\bm{H_{s_m}}\}_{i=1}^{N-1}\in \mathcal{R}^{d\times (N-1)}\\
			& \bm{P_S^{pred}} = Match(\bm{H_S}, \bm{H_M}, sigmoid)
		\end{aligned}
	\end{equation}
	For convenience and without loss of generality, we make $m = N$ which means we mask the speaker of the $\rm N_{th}$ utterance, in the following description. We construct the self-supervised target by:
	\begin{equation}
		p_{s_i}^{target}=\left\{
		\begin{array}{cl}
			1, & \text {if}\ S_i==S_N \\
			0, & \text {otherwise}
		\end{array}\right.\\
	\end{equation}
	Then binary cross entropy loss is applied here to compute the loss of this task:
	\begin{equation}
		\begin{aligned}
			\mathcal{L}_S = & -\frac{1}{N-1}\sum_{i=1}^{N-1}(p^{target}_{s_i}*log(p^{pred}_{s_i})\\
			& +(1-p^{target}_{s_i})*log(1-p^{pred}_{s_i}))
		\end{aligned}
	\end{equation}
	The gradient of $\mathcal{L}_S$ will flow backwards to refine the representations of speaker nodes so that they can decouple speaker-aware information from the original representations. After the interaction between token nodes and speaker nodes, the token nodes will gather speaker-aware information from the speaker nodes. Therefore, we denote the token representations as speaker-aware: $\bm{H_T}^s = \bm{H_T}\in \mathcal{R}^{d\times n}$, which will be forwarded to the fusion-prediction layer described in next section.

	\subsection{Fusion-Prediction Layer}
	\label{FPL}
	Given the key-utterance-aware token representation $\bm{H_T}^k$ and the speaker-aware token representations $\bm{H_T}^s$, we first fuse these two kinds of decoupled representation using the following transformation:
	\begin{equation}
		\begin{aligned}
			&\bm{H_T}^{cat} = [\bm{H_T}^k;\bm{H_T}^s;\bm{H_T}^k-\bm{H_T}^s;\bm{H_T}^k\odot\bm{H_T}^s]\\
			&\bm{H_T}^f = \text{Tanh}(W^f\bm{H_T}^{cat})\in \mathcal{R}^{d\times n}
		\end{aligned}
	\end{equation}
	where $W^f\in \mathcal{R}^{d\times 4d}$ is a linear transformation matrix with trainable weights and $\text{Tanh}$ is a non-linear activation function.\\
	Then we compute the start and end distributions over the tokens by:
	\begin{equation}
		\begin{aligned}
			& \bm{P}_{start} = \text{softmax}(\bm{w}_{start}^T\bm{H_T}^f)\odot \bm{P_U}^{expand}\\
			& \bm{P}_{end} = \text{softmax}(\bm{w}_{end}^T\bm{H_T}^f)\odot \bm{P_U}^{expand}
		\end{aligned}
	\end{equation}
	where $\bm{w}_{start}$ and $\bm{w}_{end}$ are vectors of size $\mathcal{R}^{d}$ with trainable weights, $\bm{P_U}^{expand}$ is defined on Section \ref{KIDB} and $\odot$ is element-wise multiplication.\\
	Given the ground truth label of answer span $[a_s, a_e]$, cross entropy loss is adopted to train our model:
	\begin{equation}
		\mathcal{L}_{SE} = -(\text{log}(\bm{P}_{start}[a_s])+\text{log}(\bm{P}_{end}[a_e]))
	\end{equation}
	
	If the dataset contains unanswerable question, the representation of $\bm{H_T}^f$ at $[CLS]$ position $x$ is used to predict whether a question is answerable or not:
	\begin{equation}
		p_a = \text{sigmoid}(\bm{w}^T\bm{H_T}^f[x]+\bm{b})
	\end{equation}
	where $\bm{w}^T$ and $\bm{b}$ are vectors of size $\mathcal{R}^d$ with trainable weights.\\
	Given the ground truth of answerability $t_a\in \{0,1\}$, binary cross entropy is applied to compute the answerable loss:
	\begin{equation}
		\label{AL}
		\begin{aligned}
			\mathcal{L}_A = &-((1-t_a)*log(1-p_a)\\
			&+t_a*log(p_a))
		\end{aligned}
	\end{equation}
	
	The final loss is the summation of the above losses:
	\begin{equation}
		\mathcal{L} = \mathcal{L}_U + \mathcal{L}_S + \mathcal{L}_{SE}\  (+\mathcal{L}_A)
	\end{equation}

	\section{Experiments}
	
	\subsection{Benchmark Datasets}
	We adopt FriendsQA \citep{yang2019friendsqa} and Molweni \cite{li2020molweni}, two span-based extractive dialogue MRC datasets, as the benchmarks.
	Molweni is derived from the large-scale multi-party dialogue dataset --- Ubuntu Chat Corpus \citep{lowe2015ubuntu}, whose main theme is technical discussions about problems on Ubuntu system. This dataset features in its informal speaking style and domain-specific technical terms. In total, it contains 10,000 dialogues whose average and maximum number of speakers is 3.51 and 9 respectively. Each dialogue is short in length with the average and maximum number of tokens 104.4 and 208 respectively. Unanswerable questions are asked in this dataset, hence the answerable loss in Eq. (\ref{AL}) is applied. Additionally, this dataset is equipped with discourse parsing annotations which is not used by our model however.\\
	To evaluate our model more comprehensively, another open-domain dialogue MRC dataset FriendsQA is also used to conduct our experiments. FriendsQA excerpts 1,222 scenes and 10,610 open-domain questions from the first four seasons of a well-known American TV show \emph{Friends} to tackle dialogue MRC on everyday conversations. 
	Each dialogue is longer in length and involves more speakers, resulting in more complicated speaker information flows compared to Molweni. 
	For each dialogue context, at least 4 out of 6 types (5W1H) of questions, are generated. This dataset features in its colloquial language style filled with sarcasms, metaphors, humors, etc.
	
	\subsection{Implementation Details}
	We implement our model based on \emph{Transformers} Library \citep{wolf2020transformers}. The number of information decoupling layers $L$ is chosen from 3 - 5 according to the type of the PrLM in our experiments. For Molweni, we set batch size to 8, learning rate to 1.2e-5 and maximum input sequence length of the Transformer blocks to 384. For FriendsQA, they are 4, 4e-6 and 512 respectively. Note that in FriendsQA, there are dialogue contexts whose length (in tokens) are larger than 512. We split those contexts to pieces and choose the answer with highest span probability $p_{start}* p_{end}$ as the final prediction${}^1$.
	
	\let\thefootnote\relax\footnotetext{${}^1$Codes and data are available at \url{https://github.com/EricLee8/Multi-party-Dialogue-MRC}}
	
	\subsection{Baseline Models}
	For FriendsQA, we adopt BERT as the baseline model follow \citet{li2020transformers} and \citet{liu2020graph}. For Molweni, we follow \citet{li2021dadgraph} who also employ BERT as the baseline model. In addition, we also adpot ELECTRA \citep{clark2020electra} as a strong baseline in both datasets to see if our model still holds on top of stronger PrLMs.
	
	\subsection{Results}
	Table \ref{FriendsResult} shows our experimental results on FriendsQA. BERT$_{\text{ULM+UOP}}$ \citep{li2020transformers} is a method using pretrain-fine-tune form. They first pre-train BERT on FriendsQA and additional transcripts from Seasons 5-10 of \emph{Friends} using well designed pre-training tasks Utterance-level-Masked-LM (ULM) and Utterance-Order-Prediction (UOP), then fine-tune it on dialogue MRC task. BERT$_{\text{graph}}$ \citep{liu2020graph} is a graph-based model that integrates relation knowledge and coreference knowledge using Relational Graph Convolution Networks (R-GCNs) \citep{schlichtkrull2018modeling}. Note that this model utilizes additional labeled data on coreference resolution \citep{chen2017robust} and character relation \citep{yu2020dialogue}. We adopt the same evaluation metrics as \citet{li2020molweni}: exactly match (EM) and F1 score. Our model reaches new state-of-the-art (SOTA) result on EM metric and comparable result on F1 metric, even without any additional labeled data. Besides, our model still gains great performance improvement under ELECTRA-based condition, which demonstrates the effectiveness of our model over strong PrLMs.
	\begin{table}[htbp]
		\centering
		\begin{tabular}{l c c}
			\thickhline
			Model & $\mathbf{EM}$ & $\mathbf{F1}$\\
			\hline \hline
			BERT$_{\text{basline}}$ & $43.3$ & $59.3$\\
			BERT$_{\text{ULM+UOP}}$ (\citeauthor{li2020transformers}) & $46.8$ & $63.1$\\
			BERT$_{\text{graph}}$ (\citeauthor{liu2020graph}) & $46.4$ & $\mathbf{64.3}$\\
			BERT$_{\text {our}}$ & $\mathbf{46.9}$ & $63.9$\\
			\hline
			ELECTRA$_{\text{basline}}$ & $52.8$ & $70.1$\\
			ELECTRA$_{\text{our}}$ & $\mathbf{55.8}$ & $\mathbf{72.3}$\\
			\thickhline
		\end{tabular}
		\caption{Results on FriendsQA}
		\label{FriendsResult}
	\end{table}
	
	Table \ref{MolweniResult} presents our experimental results on Molweni. Public Baseline is directly taken from the original paper of Molweni \citep{li2020molweni}. DADGraph \citep{li2021dadgraph} is the current SOTA model that utilizes Graph Convolution Network (GCN) and the additional discourse annotations in Molweni to explicitly model the discourse structure. We see from the the table that our model outperforms strong baselines and the current SOTA model by a large margin, even under the condition that we do not make use of additional discourse annotations.
	\begin{table}[htbp]
		\centering
		\begin{tabular}{l c c}
			\thickhline
			Model & $\mathbf{EM}$ & $\mathbf{F1}$\\
			\hline \hline
			BERT$_{\text{public\ basline}}$ (\citeauthor{li2020molweni}) & $45.3$ & $58.0$\\
			BERT$_{\text{our\ basline}}$ & $45.8$ & $60.2$\\
			BERT$_{\text{DADGraph}}$ (\citeauthor{li2021dadgraph}) & $46.5$ & $61.5$\\
			BERT$_{\text{our}}$ & $\mathbf{49.2}$ & $\mathbf{64.0}$\\
			\hline
			ELECTRA$_{\text{basline}}$ & $56.8$ & $70.6$\\
			ELECTRA$_{\text{our}}$ & $\mathbf{58.0}$ & $\mathbf{72.9}$\\
			\thickhline
		\end{tabular}
		\caption{Results on Molweni}
		\label{MolweniResult}
	\end{table}
	
	\section{Analysis}
	\subsection{Performance Gain Analysis}
	To get more detailed insights on our proposed method, we analyze the results on different question types of FriendsQA over ELECTRA-based model. Also, we compare our model with the baseline model on these types to see where the performance gains come from. Table \ref{PerformanceGains} shows the results of our model on different question types. Dist. means the distribution of each question type, from which we see that the question type of FriendsQA is nearly uniformly distributed. 
	
	Performance gains mainly come from question type \emph{Who}, \emph{When} and \emph{What}. We argue that the speaker information decoupling block is the predominant contributor to \emph{Who} question type since answering this type of question requires the model to have a deep understanding of speaker information flows and solve problems like coreference resolution, which is the same as our self-supervised speaker prediction task. For question type \emph{When}, the key-utterance information decoupling block contributes the most. The answer of question type \emph{When} usually comes from a scene description utterance, hence grabbing key-utterance information helps answer this kind of question. Among these improvements, question type \emph{Who} benefits the most from our model, demonstrating the strong capability of the self-supervised speaker prediction task.
	
	\begin{table}[tbp]
		\centering
		\begin{tabular}{c|c||c|c}
			\thickhline
			Type & Dist. & $\mathbf{EM}$ & $\mathbf{F1}$\\
			\hline \hline
			Who & $18.82$ & $66.8(\uparrow \mathbf{6.2})$ & $74.6(\uparrow \mathbf{4.7})$\\
			When & $13.57$ & $63.2(\uparrow \mathbf{6.1})$ & $74.1(\uparrow \mathbf{3.3})$\\
			What & $18.48$ & $58.6(\uparrow \mathbf{5.0})$ & $76.9(\uparrow 1.9)$\\
			Where & $18.16$ & $64.2(\uparrow 0.9)$ & $79.3(\uparrow 1.4)$\\
			Why & $15.65$ & $36.2(\downarrow 0.5)$ & $62.9(\uparrow 1.4)$\\
			How & $15.32$ & $41.3(\downarrow 0.9)$ & $63.5(\uparrow 0.1)$\\
			\thickhline
		\end{tabular}
		\caption{Results on different question types, where up arrows$\uparrow$ represent performance gain and down arrows$\downarrow$ represent performance drop compared to the baseline model. Significant gains (greater than 3\%) are marked as \textbf{bold}.}
		\label{PerformanceGains}
	\end{table}
	
	\subsection{Ablation Study}
	We conduct ablation study to see the contribution of each module. Table \ref{AblationResult} shows the results of our ablation study. Here KIDB and SIDB are the abbreviation of Key-utterance Information Decoupling Block and Speaker Information Decoupling Block respectively. We see from the results that both of the two modules contributes to the performance gains of our final model. For FirendsQA, SIDB contributes more and otherwise for Molweni. This is because dialogue contexts in FriendsQA tend to be long, involve more speakers and carry more complex speaker information flows. On the contrary, dialogue contexts in Molweni are short with less turns and most of the questions can be answered by considering only one key-utterance.
	
	To further investigate the effectiveness of our self-supervised speaker prediction task, we design a SpeakerEmb model in which we replace the speaker-aware token representation $\bm{H_T}^s$ by speaker representations. The speaker representations are obtained by simply gathering embeddings from a trainable embedding look-up table according to the name of the speaker. Experimental results show that it only makes a slight performance gain compared to SIDB, demonstrating that simply adding speaker information is sub-optimal compared to implicitly modeling speaker information flows using our self-supervised speaker prediction task.
	\begin{table}[tbp]
		\centering
		\begin{tabular}{lcccc}
			\thickhline \multirow{2}{*}{Model} & \multicolumn{2}{c} {FriendsQA} & \multicolumn{2}{c} {Molweni}\\
			& EM & F1 & EM & F1\\
			\hline \hline 
			Our Model & $\mathbf{55.8}$ & $\mathbf{72.3}$ & $58.0$ & $\mathbf{72.9}$\\
			\quad w/o KIDB & $55.4$ & $71.7$ & $57.7$ & $72.1$\\
			\quad w/o SIDB & $55.0$ & $71.4$ & $\mathbf{58.2}$ & $71.8$\\
			SpeakerEmb & $55.5$ & $71.9$ & $57.5$ & $72.4$\\
			\thickhline
		\end{tabular}
		\caption{Results of Ablation Study}
		\label{AblationResult}
	\end{table}
	
	\subsection{Influence of Detaching Operation}
	\label{detaching}
	We conduct experiments to investigate the influence of detaching operation mentioned in Section \ref{SIDB}. As shown in Table \ref{Detach}, if we do not detach $\bm{E}$ from the original computation graph when performing the speaker prediction task, the prediction accuracy reaches 96.8\% in the test set of FriendsQA, indicating obvious label leakage. In the meantime, the EM and F1 scores drop to 54.5\% and 70.8\%, respectively. On the contrary, our model reaches a speaker prediction accuracy of 80.8\%, which demonstrates that the detaching operation can effectively prevent label leakage.
	
	\subsection{Influence of Speaker and Utterance Numbers}
\begin{figure}[tbp]
		\centering
		\subfigure[Scores vs. Number of Speakers]{
			\begin{minipage}{0.47\textwidth} 
				\includegraphics[width=0.95\textwidth]{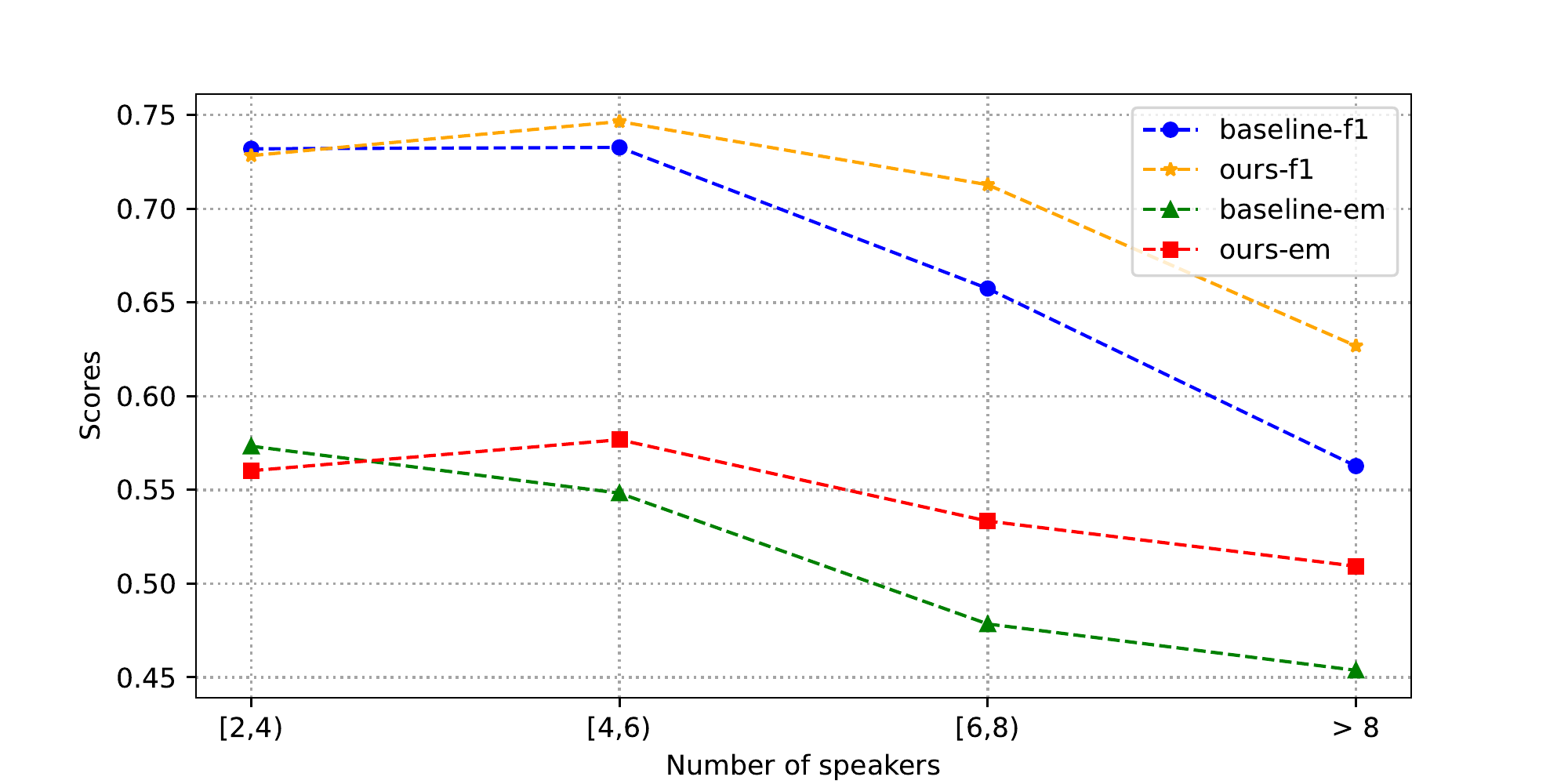} \\
			\end{minipage}
		}
		
		\subfigure[Scores vs. Number of Utterances]{
			\begin{minipage}{0.47\textwidth} 
				\includegraphics[width=0.95\textwidth]{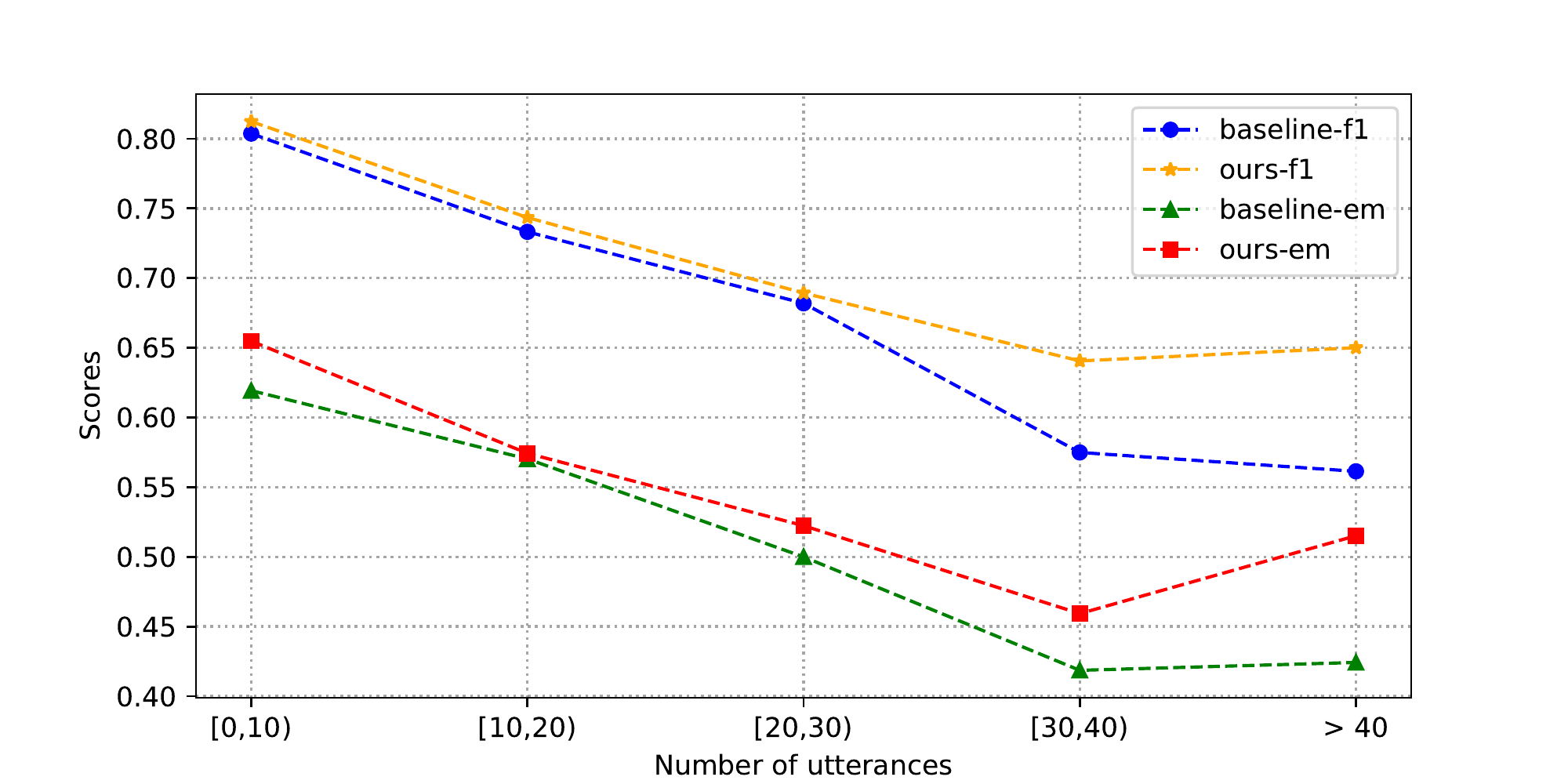} \\
			\end{minipage}
		}
		\caption{Influence of Speaker and Utterance Numbers}
		\label{number_study}
	\end{figure}

	Figure \ref{number_study} illustrates the model performance with regard to the number of speakers and utterances on FriendsQA. At the beginning, the baseline model has similar performance to our model. However, with the number of speakers and utterances increasing, there is a growing performance gap between the baseline model and our model. This observation demonstrates that our SIDB and KIDB have strong abilities to deal with more complex dialogue contexts with a larger number of speakers and utterances.
	
	\begin{table}[tbp]
		\centering
		\begin{tabular}{l c c c}
			\thickhline
			Model & $\mathbf{EM}$ & $\mathbf{F1}$ & $\mathbf{Speaker}$\\
			\hline \hline
			Our Model & $\mathbf{55.8}$ & $\mathbf{72.3}$ & $80.8$\\
			\quad w/o Detaching & 54.5 & 70.8 & 96.8\\
			\thickhline
		\end{tabular}
		\caption{Influence of Detaching Operation}
		\label{Detach}
	\end{table}
	
	\subsection{Case Study}
	To get more intuitive explanations of our model, we select two cases from FriendsQA in which the baseline model fails to answer (F1 = 0, or "exactly not match") but our model is able to answer (exactly match).
	\begin{figure}[tbp]
		\includegraphics[width=0.45\textwidth]{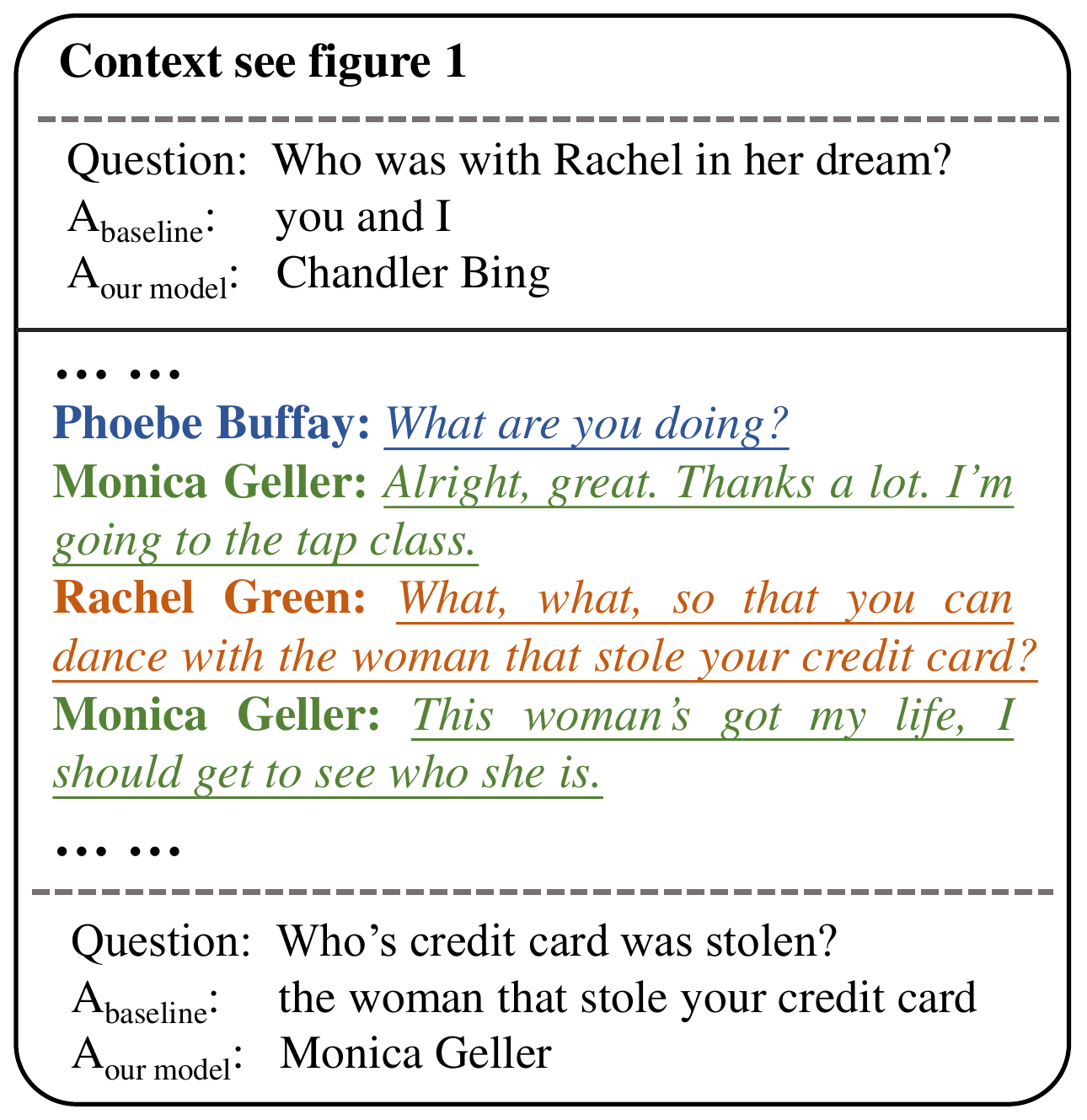}
		\centering
		\caption{Two cases from FriendsQA} 
		\label{case_study_pic}
	\end{figure}
	Figure \ref{case_study_pic} illustrates two cases where the context of the first one is shown in Figure \ref{speaker_info_flow}.
	
	In the first case, the baseline model simply predicts that \emph{"you and I"} were in \emph{Rachel's dream} while fails to notice that \emph{"you"} here refers to \emph{Chandler}. On the contrary, our model is able to capture this information since it helps the speaker prediction task. In fact, if we mask \emph{Rachel} in $\rm U_9$, our model could tell the masked speaker is \emph{Rachel}, indicating that it knows it should be \emph{Rachel} who had a dream and $\rm U_9$ is in response to $\rm U_8$.
	
	Similar observations can be seen in the second case. The baseline model simply matches the semantic meaning of the question and the context then makes a wrong prediction. Compared with the baseline model, our model has the ability to catch the information flow from \emph{Rachel} to \emph{Monica} thus predicts the answer correctly.

	\section{Conclusion}
	In this paper, for multi-party dialogue MRC, we propose two novel self- and pseudo-self-supervised prediction tasks on speaker and key-utterance to capture salient clues in a long and noisy dialogue. Experimental results on two multi-party dialogue MRC benchmarks, FriendsQA and Molweni, have justified the effectiveness of our model.

	\bibliography{main}

\begin{thebibliography}{39}
\expandafter\ifx\csname natexlab\endcsname\relax\def\natexlab#1{#1}\fi

\bibitem[{Ba et~al.(2016)Ba, Kiros, and Hinton}]{ba2016layer}
Jimmy~Lei Ba, Jamie~Ryan Kiros, and Geoffrey~E. Hinton. 2016.
\newblock \href {http://arxiv.org/abs/1607.06450} {Layer normalization}.

\bibitem[{Chen et~al.(2017)Chen, Zhou, and Choi}]{chen2017robust}
Henry~Y Chen, Ethan Zhou, and Jinho~D Choi. 2017.
\newblock \href {https://aclanthology.org/K17-1023/} {Robust coreference
  resolution and entity linking on dialogues: Character identification on tv
  show transcripts}.
\newblock In \emph{Proceedings of the 21st Conference on Computational Natural
  Language Learning (CoNLL 2017)}, pages 216--225.

\bibitem[{Clark et~al.(2020)Clark, Luong, Le, and Manning}]{clark2020electra}
Kevin Clark, Minh-Thang Luong, Quoc~V. Le, and Christopher~D. Manning. 2020.
\newblock \href {https://openreview.net/pdf?id=r1xMH1BtvB} {{ELECTRA}:
  Pre-training text encoders as discriminators rather than generators}.
\newblock In \emph{ICLR}.

\bibitem[{Cui et~al.(2020)Cui, Wu, Liu, Zhang, and Zhou}]{cui2020mutual}
Leyang Cui, Yu~Wu, Shujie Liu, Yue Zhang, and Ming Zhou. 2020.
\newblock \href {https://aclanthology.org/2020.acl-main.130/} {Mutual: A
  dataset for multi-turn dialogue reasoning}.
\newblock In \emph{Proceedings of the 58th Annual Meeting of the Association
  for Computational Linguistics}, pages 1406--1416.

\bibitem[{Devlin et~al.(2019)Devlin, Chang, Lee, and
  Toutanova}]{devlin2019bert}
Jacob Devlin, Ming-Wei Chang, Kenton Lee, and Kristina Toutanova. 2019.
\newblock \href {https://aclanthology.org/N19-1423/} {Bert: Pre-training of
  deep bidirectional transformers for language understanding}.
\newblock In \emph{Proceedings of the 2019 Conference of the North American
  Chapter of the Association for Computational Linguistics: Human Language
  Technologies, Volume 1 (Long and Short Papers)}, pages 4171--4186.

\bibitem[{Ghosal et~al.(2019)Ghosal, Majumder, Poria, Chhaya, and
  Gelbukh}]{ghosal2019dialoguegcn}
Deepanway Ghosal, Navonil Majumder, Soujanya Poria, Niyati Chhaya, and
  Alexander Gelbukh. 2019.
\newblock \href {https://aclanthology.org/D19-1015/} {Dialoguegcn: A graph
  convolutional neural network for emotion recognition in conversation}.
\newblock In \emph{Proceedings of the 2019 Conference on Empirical Methods in
  Natural Language Processing and the 9th International Joint Conference on
  Natural Language Processing (EMNLP-IJCNLP)}, pages 154--164.

\bibitem[{Gu et~al.(2020)Gu, Li, Liu, Ling, Su, Wei, and Zhu}]{gu2020speaker}
Jia-Chen Gu, Tianda Li, Quan Liu, Zhen-Hua Ling, Zhiming Su, Si~Wei, and
  Xiaodan Zhu. 2020.
\newblock \href {https://dl.acm.org/doi/abs/10.1145/3340531.3412330}
  {Speaker-aware bert for multi-turn response selection in retrieval-based
  chatbots}.
\newblock In \emph{Proceedings of the 29th ACM International Conference on
  Information \& Knowledge Management}, pages 2041--2044.

\bibitem[{He et~al.(2016)He, Zhang, Ren, and Sun}]{He_2016_CVPR}
Kaiming He, Xiangyu Zhang, Shaoqing Ren, and Jian Sun. 2016.
\newblock \href
  {https://openaccess.thecvf.com/content_cvpr_2016/html/He_Deep_Residual_Learning_CVPR_2016_paper.html}
  {Deep residual learning for image recognition}.
\newblock In \emph{Proceedings of the IEEE Conference on Computer Vision and
  Pattern Recognition (CVPR)}.

\bibitem[{Hermann et~al.(2015)Hermann, Ko{\v{c}}isk{\`y}, Grefenstette,
  Espeholt, Kay, Suleyman, and Blunsom}]{hermann2015teaching}
Karl~Moritz Hermann, Tom{\'a}{\v{s}} Ko{\v{c}}isk{\`y}, Edward Grefenstette,
  Lasse Espeholt, Will Kay, Mustafa Suleyman, and Phil Blunsom. 2015.
\newblock \href {https://openreview.net/forum?id=SJV2CwWubB} {Teaching machines
  to read and comprehend}.
\newblock In \emph{Proceedings of the 28th International Conference on Neural
  Information Processing Systems-Volume 1}, pages 1693--1701.

\bibitem[{Hu et~al.(2019)Hu, Chan, Liu, Zhao, Ma, and Yan}]{ijcai2019-696}
Wenpeng Hu, Zhangming Chan, Bing Liu, Dongyan Zhao, Jinwen Ma, and Rui Yan.
  2019.
\newblock \href {https://doi.org/10.24963/ijcai.2019/696} {Gsn: A
  graph-structured network for multi-party dialogues}.
\newblock In \emph{Proceedings of the Twenty-Eighth International Joint
  Conference on Artificial Intelligence, {IJCAI-19}}, pages 5010--5016.
  International Joint Conferences on Artificial Intelligence Organization.

\bibitem[{Jia et~al.(2020)Jia, Liu, Ren, Zhu, and Tang}]{jia2020multi}
Qi~Jia, Yizhu Liu, Siyu Ren, Kenny Zhu, and Haifeng Tang. 2020.
\newblock \href {https://aclanthology.org/2020.emnlp-main.150/} {Multi-turn
  response selection using dialogue dependency relations}.
\newblock In \emph{Proceedings of the 2020 Conference on Empirical Methods in
  Natural Language Processing (EMNLP)}, pages 1911--1920.

\bibitem[{Lai et~al.(2017)Lai, Xie, Liu, Yang, and Hovy}]{lai2017race}
Guokun Lai, Qizhe Xie, Hanxiao Liu, Yiming Yang, and Eduard Hovy. 2017.
\newblock \href {https://aclanthology.org/D17-1082/} {Race: Large-scale reading
  comprehension dataset from examinations}.
\newblock In \emph{Proceedings of the 2017 Conference on Empirical Methods in
  Natural Language Processing}, pages 785--794.

\bibitem[{Lan et~al.(2019)Lan, Chen, Goodman, Gimpel, Sharma, and
  Soricut}]{lan2019albert}
Zhenzhong Lan, Mingda Chen, Sebastian Goodman, Kevin Gimpel, Piyush Sharma, and
  Radu Soricut. 2019.
\newblock \href {https://openreview.net/forum?id=H1eA7AEtvS} {Albert: A lite
  bert for self-supervised learning of language representations}.
\newblock In \emph{International Conference on Learning Representations}.

\bibitem[{Li and Choi(2020)}]{li2020transformers}
Changmao Li and Jinho~D Choi. 2020.
\newblock \href {https://aclanthology.org/2020.acl-main.505/} {Transformers to
  learn hierarchical contexts in multiparty dialogue for span-based question
  answering}.
\newblock In \emph{Proceedings of the 58th Annual Meeting of the Association
  for Computational Linguistics}, pages 5709--5714.

\bibitem[{Li et~al.(2020)Li, Liu, Kan, Zheng, Wang, Lei, Liu, and
  Qin}]{li2020molweni}
Jiaqi Li, Ming Liu, Min-Yen Kan, Zihao Zheng, Zekun Wang, Wenqiang Lei, Ting
  Liu, and Bing Qin. 2020.
\newblock \href {https://aclanthology.org/2020.coling-main.238/} {Molweni: A
  challenge multiparty dialogues-based machine reading comprehension dataset
  with discourse structure}.
\newblock In \emph{Proceedings of the 28th International Conference on
  Computational Linguistics}, pages 2642--2652.

\bibitem[{Li et~al.(2021)Li, Liu, Zheng, Zhang, Qin, Kan, and
  Liu}]{li2021dadgraph}
Jiaqi Li, Ming Liu, Zihao Zheng, Heng Zhang, Bing Qin, Min-Yen Kan, and Ting
  Liu. 2021.
\newblock \href {https://arxiv.org/abs/2104.12377} {Dadgraph: A discourse-aware
  dialogue graph neural network for multiparty dialogue machine reading
  comprehension}.
\newblock \emph{arXiv preprint arXiv:2104.12377}.

\bibitem[{Liu et~al.(2020)Liu, Sui, Liu, and Zhao}]{liu2020graph}
Jian Liu, Dianbo Sui, Kang Liu, and Jun Zhao. 2020.
\newblock \href {https://aclanthology.org/2020.coling-main.219/} {Graph-based
  knowledge integration for question answering over dialogue}.
\newblock In \emph{Proceedings of the 28th International Conference on
  Computational Linguistics}, pages 2425--2435.

\bibitem[{Liu et~al.(2021)Liu, Zhang, , Zhao, Zhou, and Zhou}]{liu2021filling}
Longxiang Liu, Zhuosheng Zhang, , Hai Zhao, Xi~Zhou, and Xiang Zhou. 2021.
\newblock \href {https://ojs.aaai.org/index.php/AAAI/article/view/17582}
  {Filling the gap of utterance-aware and speaker-aware representation for
  multi-turn dialogue}.
\newblock In \emph{The Thirty-Fifth AAAI Conference on Artificial Intelligence
  (AAAI-21)}.

\bibitem[{Liu et~al.(2019)Liu, Ott, Goyal, Du, Joshi, Chen, Levy, Lewis,
  Zettlemoyer, and Stoyanov}]{liu2019roberta}
Yinhan Liu, Myle Ott, Naman Goyal, Jingfei Du, Mandar Joshi, Danqi Chen, Omer
  Levy, Mike Lewis, Luke Zettlemoyer, and Veselin Stoyanov. 2019.
\newblock \href {https://arxiv.org/abs/1907.11692} {Roberta: A robustly
  optimized bert pretraining approach}.
\newblock \emph{arXiv preprint arXiv:1907.11692}.

\bibitem[{Lowe et~al.(2015)Lowe, Pow, Serban, and Pineau}]{lowe2015ubuntu}
Ryan Lowe, Nissan Pow, Iulian~Vlad Serban, and Joelle Pineau. 2015.
\newblock \href {https://aclanthology.org/W15-4640/} {The ubuntu dialogue
  corpus: A large dataset for research in unstructured multi-turn dialogue
  systems}.
\newblock In \emph{Proceedings of the 16th Annual Meeting of the Special
  Interest Group on Discourse and Dialogue}, pages 285--294.

\bibitem[{Mou et~al.(2016)Mou, Men, Li, Xu, Zhang, Yan, and
  Jin}]{mou2016natural}
Lili Mou, Rui Men, Ge~Li, Yan Xu, Lu~Zhang, Rui Yan, and Zhi Jin. 2016.
\newblock \href {https://aclanthology.org/P16-2022/} {Natural language
  inference by tree-based convolution and heuristic matching}.
\newblock In \emph{Proceedings of the 54th Annual Meeting of the Association
  for Computational Linguistics (Volume 2: Short Papers)}, pages 130--136.

\bibitem[{Rajpurkar et~al.(2016)Rajpurkar, Zhang, Lopyrev, and
  Liang}]{rajpurkar2016squad}
Pranav Rajpurkar, Jian Zhang, Konstantin Lopyrev, and Percy Liang. 2016.
\newblock \href {https://aclanthology.org/D16-1264/} {Squad: 100,000+ questions
  for machine comprehension of text}.
\newblock In \emph{Proceedings of the 2016 Conference on Empirical Methods in
  Natural Language Processing}, pages 2383--2392.

\bibitem[{Reddy et~al.(2019)Reddy, Chen, and Manning}]{reddy2019coqa}
Siva Reddy, Danqi Chen, and Christopher~D Manning. 2019.
\newblock \href {https://aclanthology.org/Q19-1016} {Coqa: A conversational
  question answering challenge}.
\newblock \emph{Transactions of the Association for Computational Linguistics},
  7:249--266.

\bibitem[{Schlichtkrull et~al.(2018)Schlichtkrull, Kipf, Bloem, Van Den~Berg,
  Titov, and Welling}]{schlichtkrull2018modeling}
Michael Schlichtkrull, Thomas~N Kipf, Peter Bloem, Rianne Van Den~Berg, Ivan
  Titov, and Max Welling. 2018.
\newblock \href
  {https://link.springer.com/chapter/10.1007/978-3-319-93417-4_38} {Modeling
  relational data with graph convolutional networks}.
\newblock In \emph{European semantic web conference}, pages 593--607. Springer.

\bibitem[{Sennrich et~al.(2016)Sennrich, Haddow, and
  Birch}]{sennrich2016neural}
Rico Sennrich, Barry Haddow, and Alexandra Birch. 2016.
\newblock \href {https://aclanthology.org/P16-1162} {Neural machine translation
  of rare words with subword units}.
\newblock In \emph{Proceedings of the 54th Annual Meeting of the Association
  for Computational Linguistics (Volume 1: Long Papers)}, pages 1715--1725.

\bibitem[{Socher et~al.(2013)Socher, Perelygin, Wu, Chuang, Manning, Ng, and
  Potts}]{socher2013recursive}
Richard Socher, Alex Perelygin, Jean Wu, Jason Chuang, Christopher~D Manning,
  Andrew~Y Ng, and Christopher Potts. 2013.
\newblock \href {https://aclanthology.org/D13-1170/} {Recursive deep models for
  semantic compositionality over a sentiment treebank}.
\newblock In \emph{Proceedings of the 2013 conference on empirical methods in
  natural language processing}, pages 1631--1642.

\bibitem[{Sun et~al.(2019)Sun, Yu, Chen, Yu, Choi, and Cardie}]{sun2019dream}
Kai Sun, Dian Yu, Jianshu Chen, Dong Yu, Yejin Choi, and Claire Cardie. 2019.
\newblock \href {https://aclanthology.org/Q19-1014/} {Dream: A challenge data
  set and models for dialogue-based reading comprehension}.
\newblock \emph{Transactions of the Association for Computational Linguistics},
  7:217--231.

\bibitem[{Vaswani et~al.(2017)Vaswani, Shazeer, Parmar, Uszkoreit, Jones,
  Gomez, Kaiser, and Polosukhin}]{vaswani2017attention}
Ashish Vaswani, Noam Shazeer, Niki Parmar, Jakob Uszkoreit, Llion Jones,
  Aidan~N Gomez, {\L}ukasz Kaiser, and Illia Polosukhin. 2017.
\newblock \href {https://dl.acm.org/doi/abs/10.5555/3295222.3295349} {Attention
  is all you need}.
\newblock In \emph{Proceedings of the 31st International Conference on Neural
  Information Processing Systems}, pages 6000--6010.

\bibitem[{Veli{\v{c}}kovi{\'c} et~al.(2017)Veli{\v{c}}kovi{\'c}, Cucurull,
  Casanova, Romero, Lio, and Bengio}]{velivckovic2017graph}
Petar Veli{\v{c}}kovi{\'c}, Guillem Cucurull, Arantxa Casanova, Adriana Romero,
  Pietro Lio, and Yoshua Bengio. 2017.
\newblock \href {https://arxiv.org/abs/1710.10903} {Graph attention networks}.
\newblock \emph{arXiv preprint arXiv:1710.10903}.

\bibitem[{Wang et~al.(2019)Wang, Pruksachatkun, Nangia, Singh, Michael, Hill,
  Levy, and Bowman}]{wang2019superglue}
Alex Wang, Yada Pruksachatkun, Nikita Nangia, Amanpreet Singh, Julian Michael,
  Felix Hill, Omer Levy, and Samuel~R Bowman. 2019.
\newblock \href {https://dl.acm.org/doi/abs/10.5555/3454287.3454581}
  {Superglue: A stickier benchmark for general-purpose language understanding
  systems}.
\newblock \emph{Advances in Neural Information Processing Systems}, 32.

\bibitem[{Wang et~al.(2018)Wang, Singh, Michael, Hill, Levy, and
  Bowman}]{wang2018glue}
Alex Wang, Amanpreet Singh, Julian Michael, Felix Hill, Omer Levy, and Samuel
  Bowman. 2018.
\newblock \href {https://aclanthology.org/W18-5446/} {Glue: A multi-task
  benchmark and analysis platform for natural language understanding}.
\newblock In \emph{Proceedings of the 2018 EMNLP Workshop BlackboxNLP:
  Analyzing and Interpreting Neural Networks for NLP}, pages 353--355.

\bibitem[{Wang et~al.(2020)Wang, Hoi, and Joty}]{wang2020response}
Weishi Wang, Steven~CH Hoi, and Shafiq Joty. 2020.
\newblock \href {https://aclanthology.org/2020.emnlp-main.533/} {Response
  selection for multi-party conversations with dynamic topic tracking}.
\newblock In \emph{Proceedings of the 2020 Conference on Empirical Methods in
  Natural Language Processing (EMNLP)}, pages 6581--6591.

\bibitem[{Wolf et~al.(2020)Wolf, Chaumond, Debut, Sanh, Delangue, Moi, Cistac,
  Funtowicz, Davison, Shleifer et~al.}]{wolf2020transformers}
Thomas Wolf, Julien Chaumond, Lysandre Debut, Victor Sanh, Clement Delangue,
  Anthony Moi, Pierric Cistac, Morgan Funtowicz, Joe Davison, Sam Shleifer,
  et~al. 2020.
\newblock \href {https://aclanthology.org/volumes/2020.emnlp-demos/}
  {Transformers: State-of-the-art natural language processing}.
\newblock In \emph{Proceedings of the 2020 Conference on Empirical Methods in
  Natural Language Processing: System Demonstrations}, pages 38--45.

\bibitem[{Wu et~al.(2017)Wu, Wu, Xing, Zhou, and Li}]{wu2017sequential}
Yu~Wu, Wei Wu, Chen Xing, Ming Zhou, and Zhoujun Li. 2017.
\newblock \href {https://aclanthology.org/P17-1046/} {Sequential matching
  network: A new architecture for multi-turn response selection in
  retrieval-based chatbots}.
\newblock In \emph{Proceedings of the 55th Annual Meeting of the Association
  for Computational Linguistics (Volume 1: Long Papers)}, pages 496--505.

\bibitem[{Yang and Choi(2019)}]{yang2019friendsqa}
Zhengzhe Yang and Jinho~D Choi. 2019.
\newblock \href {https://aclanthology.org/W19-5923/} {Friendsqa: Open-domain
  question answering on tv show transcripts}.
\newblock In \emph{Proceedings of the 20th Annual SIGdial Meeting on Discourse
  and Dialogue}, pages 188--197.

\bibitem[{Yang et~al.(2019)Yang, Dai, Yang, Carbonell, Salakhutdinov, and
  Le}]{yang2019xlnet}
Zhilin Yang, Zihang Dai, Yiming Yang, Jaime Carbonell, Russ~R Salakhutdinov,
  and Quoc~V Le. 2019.
\newblock \href
  {https://proceedings.neurips.cc/paper/2019/hash/dc6a7e655d7e5840e66733e9ee67cc69-Abstract.html}
  {Xlnet: Generalized autoregressive pretraining for language understanding}.
\newblock \emph{Advances in Neural Information Processing Systems},
  32:5753--5763.

\bibitem[{Yu et~al.(2020)Yu, Sun, Cardie, and Yu}]{yu2020dialogue}
Dian Yu, Kai Sun, Claire Cardie, and Dong Yu. 2020.
\newblock \href {https://aclanthology.org/2020.acl-main.444/} {Dialogue-based
  relation extraction}.
\newblock In \emph{Proceedings of the 58th Annual Meeting of the Association
  for Computational Linguistics}, pages 4927--4940.

\bibitem[{Zhang et~al.(2018)Zhang, Li, Zhu, Zhao, and Liu}]{zhang2018modeling}
Zhuosheng Zhang, Jiangtong Li, Pengfei Zhu, Hai Zhao, and Gongshen Liu. 2018.
\newblock \href {https://aclanthology.org/C18-1317/} {Modeling multi-turn
  conversation with deep utterance aggregation}.
\newblock In \emph{Proceedings of the 27th International Conference on
  Computational Linguistics}, pages 3740--3752.

\bibitem[{Zhang et~al.(2021)Zhang, Li, and Zhao}]{zhang2021multi}
Zhuosheng Zhang, Junlong Li, and Hai Zhao. 2021.
\newblock \href {https://ieeexplore.ieee.org/abstract/document/9352490}
  {Multi-turn dialogue reading comprehension with pivot turns and knowledge}.
\newblock \emph{IEEE/ACM Transactions on Audio, Speech, and Language
  Processing}, 29:1161--1173.

\end{thebibliography}
	\bibliographystyle{acl_natbib}
	
	\appendix
	
	
	
\end{document}